\documentclass[journal,10pt]{IEEEtran}
\usepackage{cite}
\ifCLASSINFOpdf
\else
\fi
\usepackage[cmex10]{amsmath}
\usepackage{fixltx2e}
\usepackage{url}
\usepackage{times}
\usepackage{epsfig}
\usepackage{graphicx}
\usepackage{epstopdf}

\usepackage{amsmath}
\usepackage{amssymb}

\usepackage{color}
\usepackage{makecell}
\usepackage{diagbox}

\DeclareGraphicsExtensions{.eps}

\usepackage{algorithm2e}
\let\savedalgorithm\algorithm
\let\savedendalgorithm\endalgorithm

\usepackage{algorithm}

\usepackage[margin=4pt,font=footnotesize,labelfont=bf,labelsep=endash,tableposition=top]{caption}

\usepackage{footnote}
\makesavenoteenv{tabular}

\def\T{{\!\top}}

\def\x{{\boldsymbol x}}

\def\bx{ {\bf x } }

\def\bw{ {\bf w } }

\def\eg{{\rm e.g. }}
\def\ie{{\rm i.e. }}

\newcommand{\etal}{\mbox{\emph{et al.\ }}}

\definecolor{shadecolour}{gray}{0.4}

\newcommand{\COne}{{\lambda_1}}
\newcommand{\CTwo}{{\lambda_2}}
\newcommand{\Scores}{{\mathcal R}}

\newcommand{\jointPos}{{\mathbf j}}

\newcommand{\DSet}{\mathcal{D}}
\newcommand{\TrainSet}{\mathcal{X}}
\newcommand{\SimMat}{\mathbf Q}
\renewcommand{\sim}{q}
\newcommand{\score}{r}
\newcommand{\scoreVec}{{\mathbf r}}

\begin{document}
\title{Learning discriminative trajectorylet detector sets for accurate skeleton-based action recognition}

\author{Ruizhi Qiao, Lingqiao Liu, Chunhua Shen, and Anton von den Hengel%
\thanks{The authors are with
the School of Computer Science,  University of Adelaide. Adelaide, Adelaide, SA 5005, Australia.
(email: \{ruizhi.qiao, lingqiao.liu, chunhua.shen,  anton.vandenhengel\}@adelaide.edu.au)}%
\thanks{C. Shen and A. van den Hengel are also with the Australian Centre for Robotic Vision.
This work was in part supported by the Data to Decisions Cooperative Research Centre.}
\thanks{Correspondence should be addressed to C. Shen.}
}

\markboth{Discriminative trajectorylet detector sets for action recognition}%
{Qiao \MakeLowercase{\textit{et al.}}: Discriminative trajectorylet detector sets for action recognition}
\maketitle

\begin{abstract}
 The introduction of low-cost RGB-D sensors has promoted the research in skeleton-based human action recognition. Devising a representation suitable for characterising actions on the basis of noisy skeleton sequences remains a challenge, however. We here provide two insights into this challenge. First, we show that the discriminative information of a skeleton sequence usually resides in a short temporal interval and we propose a simple-but-effective local descriptor called trajectorylet to capture the static and kinematic information within this interval. Second, we further propose to encode each trajectorylet with a discriminative trajectorylet detector set which is selected from a large number of candidate detectors trained through exemplar-SVMs. The action-level representation is obtained by pooling trajectorylet encodings. Evaluating on standard datasets acquired from the Kinect sensor, it is demonstrated that our method obtains superior results over existing approaches under various experimental setups.
\end{abstract}

\begin{IEEEkeywords}
Action recognition, Kinect sensor, exemplar support vector machines, feature learning, 3D action feature representation.
\end{IEEEkeywords}

\IEEEpeerreviewmaketitle

\section{Introduction}
\IEEEPARstart{T}{he} recognition of human actions is an active research field in recent years and much effort has been made to address this problem \cite{action_survey08}. Intuitively, a temporal sequence of 3D skeleton joint locations captures sufficient information to distinguish between actions, but recording skeleton sequence was very expensive with the traditional motion capture technology,  which limits the applications to which it has been applied \cite{MoCap_review06}. Recently, with the advent of RGB-D cameras such as Microsoft Kinect \cite{Kinect_intro13}, the acquisition of 3D skeleton data for action recognition has become much easier and faster \cite{Realtime_kinect11}. This advance promotes a number of skeleton-based action recognition approaches \cite{LieGroup14, HOD13, DBM_HMM14}. The key challenge of these approaches is how to extract discriminative features from the noisy temporally evolving skeletons.

\begin{figure}[t]
\begin{center}

   \includegraphics[width=1.0\linewidth]{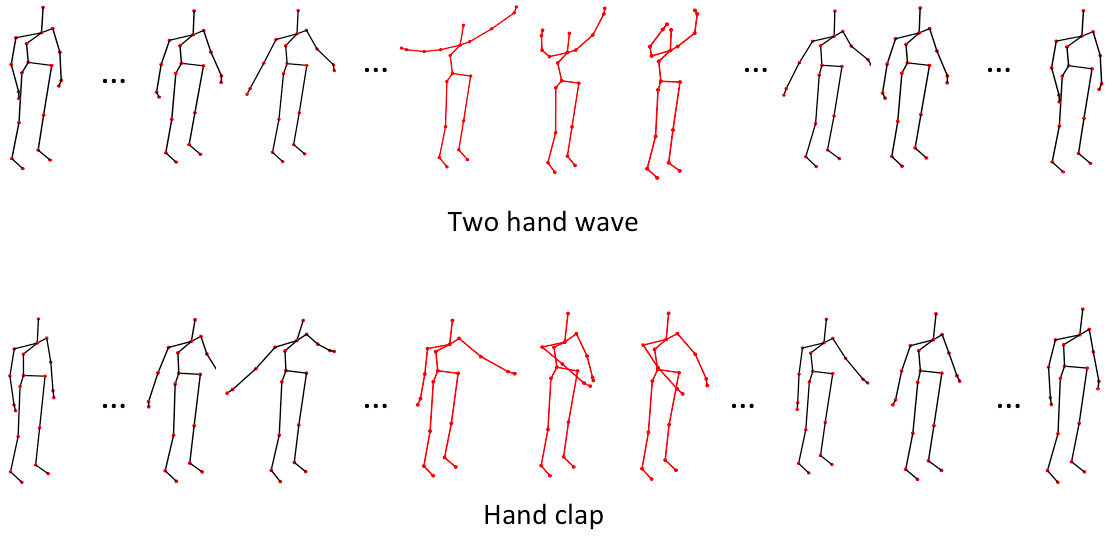}
\end{center}
   \caption{Skeleton sequences from two action classes.
   Only the red skeletons show significant differences between the two sequences. In this example, less than $20\%$ of the frames are required to tell whether the skeleton is clapping or waving.}
\label{fig: ex_temp}
\end{figure}

The trajectory of skeletal joints in space-time is the direct representation of human actions.
 Earlier works \cite{traj_intro08, traj_intro10} model human action trajectory descriptors of variable-lengths and classify them based on the similarity matching of trajectories. In \cite{HOD13}, an action representation is encoded with a histogram voted by the displacements of joint trajectories with respect to their orientations. In these works, the global feature is extracted from the whole trajectory. However, only a short section of the trajectory is actually distinctive and can provide usable information about the action being undertaken. For example, as illustrated in Figure~\ref{fig: ex_temp}, only moments when the subject moves its hands,  during the two actions of waving and clapping, are indicative of the performed action class, while all the remaining poses are irrelevant and potentially distracting. The abundant non-informative local patterns may cause large variance to the global trajectory. Compared to the global representation, later works \cite{eigenjoints12, MovingPose} explore discriminative patterns to create local descriptors at frame-level. Despite its robustness, a frame-level descriptor, without additional temporal information, hardly depicts the movement of actions, and is insufficient for recognition.

Different from the above-mentioned approaches which either represent an action with the whole sequence or extract local features at the frame level, we argue that the discriminative information of an action is better captured by a short interval of trajectories. This interval usually consists of several frames. In other words, its temporal range is longer than a single frame but much shorter than the whole skeleton sequence. To extract features from the trajectory interval, we make our first contribution by designing a novel local descriptor called trajectorylet to capture the static and dynamic motion information within the short interval.

 Furthermore, as we have observed, not all trajectorylets in a sequence are equally important for classification and the recognition performance generally benefits from focusing on the discriminative ones.
In skeleton-based action recognition, recent works \cite{MovingPose, Canonical13} directly learn the discriminative frames from the training set.  Unlike the aforementioned works, our approach does not explicitly look for the discriminative trajectorylets, but rather provides a method for creating a set of detectors that fire on specific template trajectorylets. Our approach firstly applies exemplar-SVM \cite{ESVM11} to learn a large number of candidate detectors and then selects detectors according to their discriminative performance over the trajectorylets in the training set. We further cluster detectors into multiple clusters, and remove the redundancy of the learned detectors by selecting one representative detector from each cluster. The selected detectors form a template detector set and their detection scores on a trajectorylet is utilized as the coding vector of that trajectorylet. The action level representation is then obtained by pooling all trajectorylet coding vectors and temporal pyramid pooling can also be incorporated to capture the long range temporal information of the action sequence. In extensive experiments, this framework brings significant performance improvement over state-of-the-art approaches for skeleton-based action recognition.

In summary,
our {\em first contribution}
is the trajectorylet, a novel local descriptor that captures static and dynamic information
in a short interval of joint trajectories. In our {\em second contribution},
a novel framework is proposed to generate robust and discriminative representation for action instances from a set of learned template trajectorylet detectors.

Following  briefly reviewing related literature in Section~\ref{sec: rel_work},
we propose the design of our local feature and detector learning method in Section~\ref{sec: FL}.
We then present the action-level representation of an action instance in Section~\ref{sec: GL}.
Our framework is experimentally evaluated in Section~\ref{sec: EX} and summarized in Section~\ref{sec: CONC}.

\section{Related work}
\label{sec: rel_work}
The key challenge of skeleton-based action recognition is how to construct the action representation from a sequence of skeletal joints. Some video-based methods \cite{Track_points09, dense_traj11} extract trajectories of multiple tracking points, and compute descriptors along them, such as HOG, HOF and MBH. For skeleton-based methods, trajectories are directly obtained from the space-time evolution of skeletal joints. The most straightforward way is to model the trajectory holistically, either by extracting statistics from the sequence or modelling its generative process. In \cite{HOD13}, a histogram records the displacements of joint orientations over the whole trajectory. In \cite{jointangles13}, the action is modelled with the pairwise affinities trajectories of joint angles. In \cite{HOJ3D13}, the action sequence is modelled by the Hidden Markov Model with quantized histogram of spherical coordinates of joint locations as frame-level feature. In \cite{LieGroup14}, 3D geometric relationships between various body parts are modelled with a Lie group to represent the whole action.

Besides directly modelling the trajectory holistically, it has also been noted that only a small fraction of patterns of a skeletal sequence are actually distinctive and thus many approaches have been proposed to identify those discriminative patterns, whether these patterns are defined spatially or temporally.

It has been found that not all skeletal joints are informative for distinguishing one action from the others, therefore it is beneficial to select a subset of joints. Ofli \etal \cite{SMIJ12} select a subset of most informative joints according to criteria such as mean or variance of joint angles. In \cite{Actionlet12},  joints are grouped into actionlets, and the most discriminative collection of them are mined via the multiple kernel learning approach. In \cite{bio13}, a subset of joints within a short-time interval is extracted according to the spatio-temporal hierarchy of the moving skeleton, and a linear combination of them is learned via a discriminative metric learning approach. In \cite{Pose_based13}, the distinctive set of body parts are mined from their co-occurring spatial and temporal  configurations. In \cite{Evo14}, an evolution algorithm is employed to select an optimal subset of joints for action representation and classification is performed by using DTW-based sequence matching.

As most of the frames in an action sequence are comprised of non-distinctive static poses, features at a few discriminative temporal locations are informative enough to represent an action. In video-based action recognition, a number of key frame selection approaches have been proposed. In \cite{infoframe08}, key frames are selected by ranking the conditional entropy of the codewords assigned to the frames. In \cite{Poseltekeyframe13}, the locations of key frames are modelled as latent variables and estimated for each action instance by dynamic programming. In recent works on skeleton-based action recognition, distinctive canonical poses \cite{Canonical13} are learned via logistic regression, and discriminative frames \cite{MovingPose} are identified by their approximated confidence of belonging to a specific action class. In \cite{eigenjoints12}, distinctiveness of each frame is calculated by a measurement of accumulated motion energy.

\section{The proposed action representation}
\label{sec: FL}
Our model utilizes the relationships between the positions of the $J$ skeletal joints $\jointPos_j=(x_j, y_j, z_j) \in \mathbb{R}^3, j=1 \cdots J$ in the current and preceding frames to form a local trajectorylet. Because human skeleton size varies from different action instances, we perform a skeleton size normalization on the raw skeletal joints according to \cite{MovingPose}. We also subtract the position of the hip center $\jointPos_{hip}$ from each joint and concatenate them to form a feature column: $\jointPos=[\jointPos_1 - \jointPos_{hip}, \cdots, \jointPos_J - \jointPos_{hip}] \in \mathbb{R}^{3J}$, making $\jointPos_{hip}$ the origin point of the coordinate system across all frames and subjects.

\subsection{Trajectorylet}
\begin{figure}[t]
\begin{center}

   \includegraphics[width=0.6\linewidth]{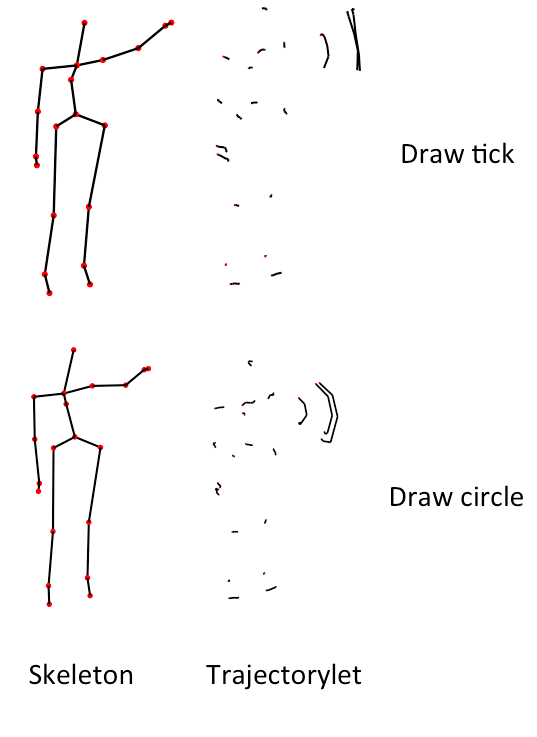}
\end{center}
   \caption{The joint coordinate information at frame-level may provide little information to distinguish between some action classes, such as the above drawing actions. One of the advantages of trajectorylets is their ability to focus on the dynamics of distinctive sections of individual actions.}
	\label{fig: visual_traj}
\end{figure}

Although holistic trajectories of joints depict the movement of human body, distinctive patterns are usually overwhelmed by common ones. For example, in long-term actions such as draw circle and draw tick, only the last moment of drawing movement distinguishes them, before which both trajectories share the same movement of raising up hand for a long time. On the other hand, as depicted in Figure~\ref{fig: visual_traj},  frame-level local descriptors record current poses and some local dynamics, but they fail to capture the movement that spans a long temporal range. To distinguish walk from run, for instance, we need to examine the displacement and speed of the joints within a sufficient period of time, rather than the static poses. Based on these observations, we propose our trajectorylet local descriptor, which captures the static and dynamic information of trajectories in a short period of time. Compared with frame-level descriptors, trajectorylet depicts richer dynamic information. On the other hand, its temporal range is much smaller than the whole trajectory sequence and therefore it is less affected by potentially irrelevant frames.

More specifically, considering a trajectorylet of length $L$ starting from frame $t_0$, we extract the static positions of the joints from each frame occurring before time $t_0 + L$:

\begin{equation}
            \label{EQ:curent_frame}
    \bx_{0}^{t_0} = [ {\jointPos^{t_0}}^\T, {\jointPos^{t_0+1}}^\T, \cdots, {\jointPos^{t_0+L-1}}^\T ]^\T \in  \mathbb{R}^{(L \times 3J)}.
        \end{equation}

In order to retrieve the dynamic information within this interval, we inspect multiple levels of temporal dynamics such as displacement and velocity.

\begin{align}
            \label{EQ:adjacent_frame1}
         & \bx_{1}^{t_0} = [ {\Delta\jointPos^{t_0+1}}^\T, \cdots, {\Delta\jointPos^{t_0+L-1}}^\T ]^\T \in  \mathbb{R}^{((L-1) \times 3J)}, \\ \notag
         & \Delta\jointPos^{t_0+i}= \jointPos^{t_0+i} -\jointPos^{t_0} ,      \, i=1, \cdots, L-1. \\
         \label{EQ:adjacent_frame2}
         & \bx_{2}^{t_0} = [ {\Delta^2\jointPos^{t_0+2}}^\T, \cdots, {\Delta^2\jointPos^{t_0+L-1}}^\T ]^\T \in  \mathbb{R}^{((L-2) \times 3J)}, \\
       & \Delta^2\jointPos^{t_0+i}=\Delta\jointPos^{t_0+i} - \Delta\jointPos^{t_0+i-1}, \, i=2, \cdots, L-1. \notag
\end{align}
where $\Delta\jointPos^{t_0+i}$ indicates the relative joint displacements of frame $t_0+i$ from the first frame;  $\Delta^2\jointPos^{t_0+i}$ indicates the joint velocities of frame $t_0+i$ from its previous frame within the trajectorylet. The static positions of ${\bx_{0}^{t}}$ store the absolute spatial location of the trajectorylet. The temporal dynamics ${\bx_{1}^{t}}$ and ${\bx_{2}^{t}}$ approximate the relatively kinematic evolution within this short time interval. Combining both static and dynamic information we define the $t$-th trajectorylet for an action instance with $F$ frames as

\begin{align}
            \label{EQ:def_traj}
    \bx^{(t)} = ({\bx_{0}^{t}}^\T,  {\bx_{1}^{t}}^\T, {\bx_{2}^{t}}^\T)^{\T} \in {\mathbb{R}}^{(3L-3)3J}.
\end{align}
where $t=1, \cdots, F-L$.

\begin{figure}[t]
\begin{center}
\begin{tabular}{c}
   \includegraphics[width=0.9\linewidth]{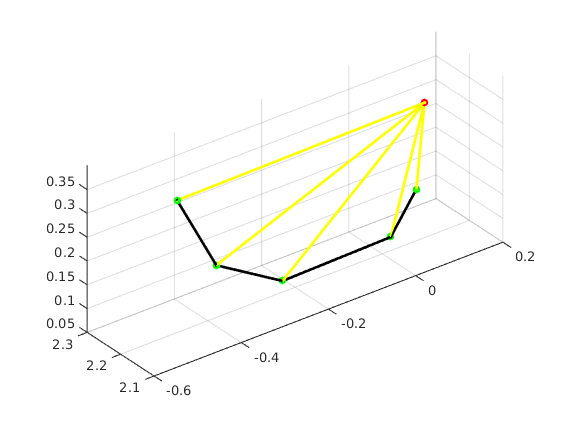} \\
   \includegraphics[width=0.9\linewidth]{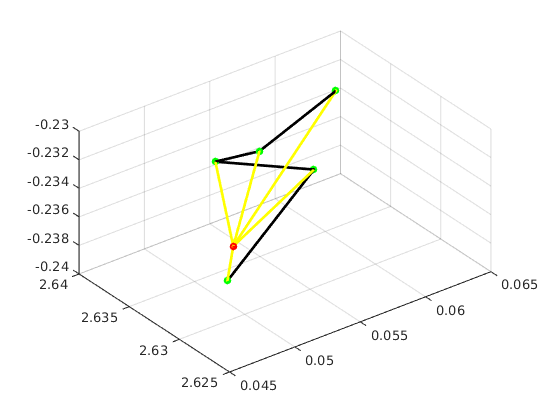}
\end{tabular}
\end{center}
   \caption{Visualization of trajetorylet of length 5 at a single joint (left hand). The red point is the position at the starting frame, and the green points are its positions at succeeding frames in this interval. The yellow segments are joint displacements  from the first frame. The black segments are joint velocities at each frame. The top trajetorylet is part of {\it drawing circle} and the bottom trajetorylet is part of {\it high waving}. The differences between them are clearly distinguished by their positions, displacements and velocities over a short period of time.}
	\label{fig: ex_traj}
\end{figure}

PCA is applied on trajectorylets to reduce the their dimension for our detector learning module. We still denote the final descriptor as $\bx^{(t)} \in {\mathbb{R}}^{d}, \, d \leq (3L-3)3J$. Figure \ref{fig: ex_traj} visualizes components in a trajectorylet , including one static component and two dynamic components.

\subsection{Learning candidate detectors of discriminative trajetorylet using ESVM}
As we have previously discussed, only a small fraction of the trajectorylets from an sequence contains sufficient information for identifying the associated action. Most of the trajectorylets, especially those that contain the static posture, are shared by multiple action classes. Our aim is to learn a set of detectors that fire on the distinctive trajectorylets. To this end, we firstly resort to exemplar-SVM (ESVM) \cite{ESVM11} to learn a large set of detectors for a large number of sampled trajectorylets, one for each sampled trajectorylet. Then for each action instance we select a few discriminative trajectorylet detectors as the candidate detectors of discriminative trajectorylet.

An ESVM learns a decision boundary that achieves largest margin between an exemplar sample and a set of negative examples.  If we take each trajectorylet as a positive exemplar $\bx_E$ of its associated class $c$, $c=1, \cdots C$, and trajectorylets that belong to other action classes as the negative examples, we can train an exemplar-SVM for it and formally this can be formulated as:

 \begin{align}
            \label{EQ:ESVM}
	\arg \min_{\bw_E,b_E}	||\bw_E||^2 + \lambda_1 h( \bw_E^\T \bx_E + b_E)  \\ \notag
	+ \lambda_2 \sum_{\bx \in {\cal N}_c} h(-\bw_E^\T \bx - b_E)
   \end{align}

where $h(x)=\max(0, 1-x)$ is the hinge loss function, and ${\cal N}_c$ is the negative set of trajectorylets that do not belong to class $c$. $\COne$ and $\CTwo$ denote the weights of loss for positive and negative samples respectively, and $\COne > \CTwo$ ensures that a greater penalty will be applied to the incorrectly classified positive exemplars.

For each ESVM, the trained detector $f(\bx)=\bw_E^\T \bx + b_E$ returns higher scores on trajectorylets that are most similar to $\bx_E$.
If the current exemplar trajectorylet is common in multiple action classes, the returned trajectorylets are abundant in multiple classes. On the contrary, if the current exemplar trajectorylet is unique for a single class, most returned trajectorylets belong to the same class with the current exemplar trajectorylet. Thus we can employ the distribution of action classes of the returned trajectorylets to estimate the discriminative power of one detector.

\begin{figure*}[ht]

\includegraphics[width=1.0\linewidth]{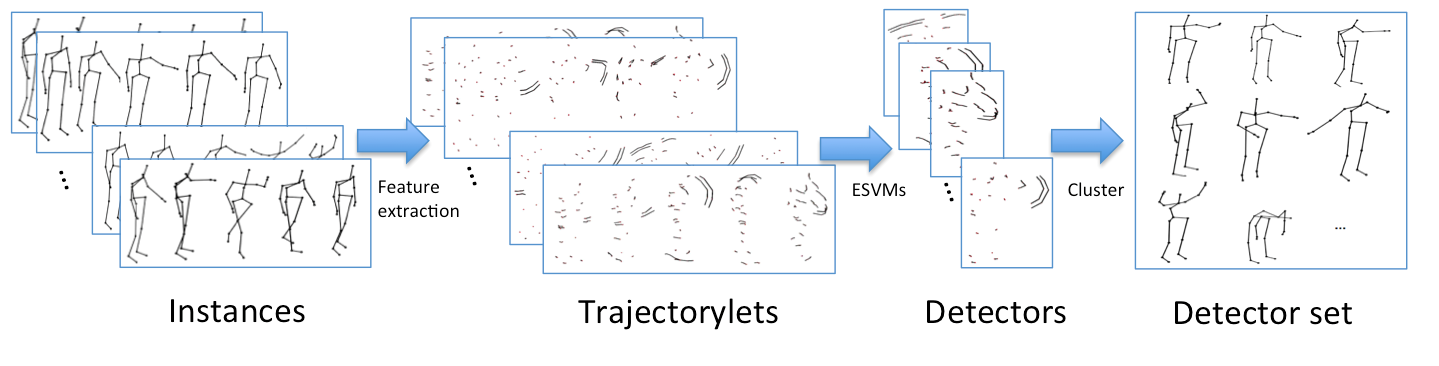}
\centering
   \caption{Overview of our feature learning framework.}
\label{fig:short}
\end{figure*}

Given an action instance $A$, we extract $F_A$ trajectorylet descriptors $\bx^{(t)}~ ,~ t=1, \cdots F_A$, and train the associated detectors $(\bw_E^{(t)}, b_E^{(t)})~,~ t=1, \cdots F_A$. A selection method is implemented to find the most discriminative trajectorylet detector among the candidates. More specifically, we apply each detector $(\bw_E^{(t)}, b_E^{(t)})$ to the trajectorylets $\bx^{(i)}$, $i=1, \cdots N$ sampled from the \textit{whole training set} and compute the detection scores $\score_{ti}=\bw_E^{(t)\T} \bx^{(i)} + b_E^{(t)}$ \footnote{In order to measure the scores on the same scale, we adjust the trained parameters with unit norm before computing the scores.}. From $\Scores_{t}=\{\score_{ti}\}_{i=1, \cdots N}$ we choose a subset $\Scores'_t$ , with the top $N_A$ scores, corresponding to the trajectorylets that are most compatible with current detector $(\bw_E^{(t)}, b_E^{(t)})$. For the $N_A$ trajectorylets detected by $(\bw_E^{(t)}, b_E^{(t)})$, we denote $h^{(c)}_t$ as the number of trajectorylets belonging to action class $c$ . The histogram $H_t=[h^{(1)}_t, \cdots, h^{(C)}_t]^{\T} \in \mathbb{R}^C$ gives a clear view of the distinctiveness of detector $(\bw_E^{(t)}, b_E^{(t)})$.

If $H_t$ is flat across many classes, $\bx^{(t)}$ is a common pattern shared by many classes and its detector is therefore not distinctive. If the $H_t$ is centered mostly at the correct class, trajectorylet $\bx^{(t)}$ is a distinctive pattern for this class and hence $(\bw_E^{(t)}, b_E^{(t)})$ is an effective detector of this distinctive pattern.
In practice, if the correct class corresponding to $(\bw_E^{(t)}, b_E^{(t)})$ is $c$, we denote $P_t=h^{(c)}_t/N_A$ as the ratio of correctly detected trajectorylets and a detector with higher $P_t$ is selected because it fires primarily on trajectorylets with the same class of it, verifying the distinctiveness of this detector. We summarize this approach in Algorithm \ref{ALG:1}.
\def\ADot{ { $\bf \cdot$ } }%

\setcounter{AlgoLine}{1}
\begin{algorithm}[ht]
\caption{\footnotesize  Find discriminative detectors for an action instance}
\centering   
{\footnotesize
   \begin{minipage}[]{0.94 \linewidth}
    \KwIn{
    Training action instance $A$ of class $c$, trajectorylets within it $\{\bx^{(t)}\}_{t = 1 \cdots F_A}$;
     sampled training trajectorylets $\TrainSet=\{ \bx^{(i)} \}_{i=1, \cdots N}$; number of trajectorylets to retain: $N_A$;
     maximum number of detectors to be selected for the instance: $M_A$. 
    }   
   { {\bf Initialize}:
       Set of discriminative detectors for instance $A$: ${\DSet_A}= \O$; number of discriminative detectors selected for the instance $m_A=0$.
   }
   
   \For { $t = 1 \cdots F_A$  }
   {
    \ADot
    		Solve ESVM  $\rightarrow (\bw_E^{(t)}, b_E^{(t)})$.

    \ADot
        Compute detection scores on sampled trajectorylet set %

    \ADot
        Compute $H_t$ from the top $N_A$ scored samples.

    \ADot
        Compute the ratio of correctness $P_t$ of $H_t$.

   }

   \ADot
        Sort $P_t$ by magnitude, storing the resulting (sorted) indeces in ${\mathbf s}$.

	\For{$t$ in ${\mathbf s}$}
        {
           \ADot
              ${\DSet_A} = {\DSet_A} \cup (\bw_E^{(t)}, b_E^{(t)})$.

           \ADot
              $m_A=m_A+1$.

        \If{$m_A \geq M_A$}
        {
           \ADot
              Break.
        }

        }

\KwOut{
     Discriminative detectors for instance $A$: ${\DSet_A}$.
}
\end{minipage}
}
\label{ALG:1}
\end{algorithm}

\subsection{Template detector set}

\label{subsec:Cluster}
As the detectors are discovered from every action instance, the size of the detector set grows with the number of training instances, which will lead to a very high-dimensional action representation and make the computation intractable. On the other hand, the above method might select similar distinctive detectors multiple times, resulting in a highly redundant detector set. To control the size of detector set and remove the redundancy of candidate detector set, we perform spectral clustering on candidate detectors and then select one detector from each cluster as the final detector set used for trajectorylet encoding. To build the affinity graph for spectral clustering, we need to specify the similarity measurement between two detectors. Here we measure this similarity by considering the ``active detection scores'' of two detectors which refer to the detection scores with positive values. We evaluate it by firstly calculating detection scores on $N$ sampled trajectorylets and setting negative detection scores to zero. This process gives a $N$ dimensional active detection score vector $\scoreVec_{d}$ for each detector and the similarity between two detectors are measured as follows:
 \begin{equation}
            \label{EQ:Sim_Mat}
		\sim_{dd'}= \frac{\scoreVec_{d}^\T \scoreVec_{d'}}{||\scoreVec_d|| \cdot ||\scoreVec_{d'}||}
        \end{equation}
where $\|\cdot\|$ represents the $l^2$ norm, and $\scoreVec_{d}$ and $\scoreVec_{d'}'$ denote the active detection score vectors for the two compared detectors. The value $\sim_{dd'}$ measures the similarity between two detectors and is used to build the affinity matrix $\SimMat$ for the detector set $\DSet$, that is,  $\SimMat = [\sim_{dd'}]_{d,d'=1,\ldots,D}$. We apply spectral clustering to $\SimMat$ and obtain $K<D$ clusters of detectors. The detectors within the same cluster fire on similar trajectorylets. From each cluster, we select a representative detector that produces the highest score on the sampled trajectorylets. In practice, given a sufficient large $K$, the collection of representative detectors can cover all discriminative trajectorylets. We call this collection the template trajectorylet detector set.

\section{Global descriptor and classification}
\label{sec: GL}
\label{sec: GL}
For the detectors in the template detector set, we evaluate their detection scores on each trajectorylet and max-pool those detection scores to obtain the action representation. Formally, let $\bx_i^j \in \mathbb{R}^n$ be the $j$-th trajectorylet of the $i$-th action, and $(\bw_k, b_k)$ be the $k$-th detector in the template detector set.
We define the action representation for the $i$-th action ${\bf \Phi}(\bx_i) = [\Phi_k(\bx_i)]_{k=1,\ldots,K} $ as:

\begin{equation}
	\label{proj_encoding}
	{ \Phi}_k(\bx_i) = \max_j {(\bw_k^\T \bx_i^j + b_k)}, \; k=1, \cdots, K.
\end{equation}

We use a one-versus-all SVM to classify actions among the $C$ action classes $y_i \in \{1, \cdots, C \}$.

The learned feature mapping ${\bf \Phi}(\cdot)$ governed by the template detector set serves as a global descriptor of the action instance. It maps temporally continuous trajectorylets into a higher-level representation. Also, ${\bf \Phi}(\cdot)$ can not only map a complete sequence of action, but also works for a temporal sub-sequence. This allows us to build a temporal pyramid representation of the action instance. For a 3-level temporal pyramid, the sub-sequences are $F^{(p)}, p=1, \cdots, 7$, and the $k$-th dimension of subfeature ${\bf \Phi}^{(p)}(\bx_i)$ for sub-sequence $p$ is

\begin{equation}
	\label{proj_encoding_TP}
	{ \Phi}^{(p)}_k(\bx_i) = \max_{j \in F^{(p)}} {(\bw_k^\T \bx_i^j + b_k)}.
\end{equation}

The concatenated ${\bf \Psi}(\cdot) = [{\bf \Phi}^{(1)}(\cdot)^\T, \cdots, {\bf \Phi}^{(7)}(\cdot)^\T]^\T$ incorporates the temporal information of the skeleton sequence. Therefore we are able to train a one-versus-all SVM with this feature that takes into account the global temporal information of the whole action sequence.

\section{Experiments}
\label{sec: EX}
We organize the experimental evaluation in four parts. We first compare our proposed method against other state-of-the-art methods on two standard datasets obtained from the Kinect sensor. Then we analyze the performance of our method under different parameter settings. Since our method consists of two modules, the trajectorylet descriptor and the template detector learning based middle-level feature representation, we conduct two experiments to separately evaluate their impact on the classification performance. To examine the first module, we compare our descriptor against the descriptor of \cite{MovingPose}, which is most related to our trajectorylet descriptor, by keeping the other settings of the recognition system the same. We also compare our descriptor with its several alternative variants.
To examine the second module, we compare our method with alternative way to obtain constructed from three state-of-the-art middle-level feature representation methods: VLAD \cite{VLAD10}, LSC \cite{LSC11}, and LLC \cite{LLC10}.

{\bf Implementation details:} The ESVMs are implemented by liblinear \cite{liblinear08}, which produces about 5 candidate detectors per second on an Intel Core i7 CPU at 3.40GHz. We set the regularization parameters as $\COne=10$ and $\CTwo=0.01$ for all ESVMs. There are on average $9,000$ and $30,000$ local descriptors in the negative sets for MSR Action3D and MSR DailyActivity3D respectively. The dimensionality of trajectorylets is reduced to 50\% percent of it by PCA. As the testing data will not be known in advance, the PCA coefficients $\mu$ and covariance matrix are learned from the training data only. Unless indicated otherwise, the length of trajectory descriptor is set to $L=5$. The regularization parameter for the final one-versus-all SVM is determined by a five-fold cross-validation. We apply a 3-level temporal pyramid on MSR DailyActivity3D only, because it contains complex actions which involves several sub-actions and the long-range temporal information can be useful in such a case.

\subsection{MSR Action3D}

\begin{table}
\begin{center}
\begin{tabular}{|l|l|l|}
\hline
AS1  & AS2 & AS3 \\
\hline\hline
 Horizontal arm wave & High arm wave & High throw  \\
Hammer & Hand catch & Forward kick  \\
Forward punch & Draw x & Side kick  \\
High throw & Draw tick & Jogging \\
Hand clap & Draw circle & Tennis swing \\
Bend & Two hand wave & Tennis serve \\
Tennis serve & Forward kick & Golf swing \\
Pickup \& throw & Side boxing & Pickup \& throw \\

\hline
\end{tabular}
\end{center}
\caption{The classes in the three action subsets of the MSR Action3D dataset.}
\label{tab: action_sets}
\end{table}

\begin{table}

\begin{center}
\begin{tabular}{|l|c|c|c|c|}

\hline
Protocol of \cite{3Dbag} & AS1  & AS2 & AS3 & Average\\
\hline\hline
3DBag \cite{3Dbag} & 72.9 & 71.9 & 79.2 & 74.7 \\
HO3DJ \cite{HOJ3D13} & 88.0 & 85.5 & 63.5 & 79.0 \\
EigenJoints \cite{eigenjoints12} & 74.5 & 76.1 & 96.4 & 82.3 \\
HOD \cite{HOD13} & 92.4  & 90.1 & 91.4 & 91.2 \\
Lie Group \cite{LieGroup14} & 95.3 & 83.9 & 98.2 & 92.5  \\
EJS \cite{Evo14} & 91.6 & 90.8 & 97.3 & 93.2  \\
Moving Pose \cite{MovingPose} \footnotemark[2]   & $\mathbf{96.4}$ &91.6 &99.1 &95.7 \\

Ours & $\mathbf{96.4}$ & $\mathbf{97.5}$ & $\mathbf{100.0}$ & $\mathbf{97.9}$ \\
\hline
\end{tabular}
\end{center}
\caption{Results on 3 subsets of the MSR Action3D dataset.}
\label{tab: action_all}
\end{table}

 \footnotetext[2]{We use the code of \cite{MovingPose} to obtain this result, as the original work did not report the results according to the protocol of \cite{3Dbag}.}

The MSR Action3D dataset consists of human actions expressed with skeletons composed of 20 3D body joint positions in each frame. The 20 joints are connected by 19 limbs. There are 20 action classes performed by 10 subjects for 2 or 3 times each, making up 567 action instances. Each action instance contains a temporal sequence of a moving skeleton, usually in 30-50 frames. As in \cite{Actionlet12} and \cite{MovingPose}, we drop 10 instances because they contain erroneous data. The experiment setup is that of a cross-subject test \cite{3Dbag}, \ie  instances of half of the subjects are used for training and instances of the other half subjects are used for testing. We construct $H_t$ with top responding $N_A=50$ trajectorylets, and select $M_A=10$ best detectors for each training instance. We use the clustering method of section~\ref{subsec:Cluster} to obtain the template trajectorylet detector set. The final number of template detectors is set to $K=500$.

In Table~\ref{tab: action_all}, we compare our approach with other state-of-the-art methods using the protocol of \cite{3Dbag}, by which the 20 action classes are grouped into 3 action subsets AS1, AS2, and AS3. The training and testing is performed on each action set separately. AS1 and AS2 group actions with similar movements while AS3 group complex actions. The action classes of each action subset are listed in Table~\ref{tab: action_sets}. On average, our proposed method is more accurate than all other methods. On AS2, all other methods get moderate accuracy and in contrast our method outperforms the second best by 5.9\%. Note that on AS3, our method achieves perfect recognition.

\begin{table}
\begin{center}
\begin{tabular}{|l|c|}
\hline
Protocol of \cite{Actionlet12} & Accuracy \\
\hline\hline
Recurrent Neural Network \cite{RNN11} & 42.5 \\
Dynamic Temporal Warping \cite{DTW06} & 54.0 \\
Canonical Poses \cite{Canonical13} & 65.7 \\
DBM+HMM  \cite{DBM_HMM14} & 82.0 \\
JAS (skeleton data only) \cite{jointangles13} & 83.5 \\
Actionlet Ensemble \cite{Actionlet12} & 88.2 \\
HON4D \cite{HON4D} & 88.9 \\
Lie Group \cite{LieGroup14} & 89.5  \\
LDS   \cite{bio13} \footnotemark[3]      & 90.0     \\
Pose based \cite{Pose_based13} & 90.2 \\
Moving Pose \cite{MovingPose}  & 91.7 \\
Ours & $\mathbf{95.9}$\\
\hline
\end{tabular}
\end{center}
\caption{Results on the entire MSR Action3D dataset.}
\label{tab: action}
\end{table}

 \footnotetext[3]{ It should be noted that the result of \cite{bio13} here is not obtained under the same setting as ours. This approach selected a subset of 17 actions
performed by 8 subjects, 5 for training and 3 for testing, consisting of 379 action instances in total.}

\begin{figure*}[h]
\begin{center}
  \includegraphics[width=0.8\linewidth]{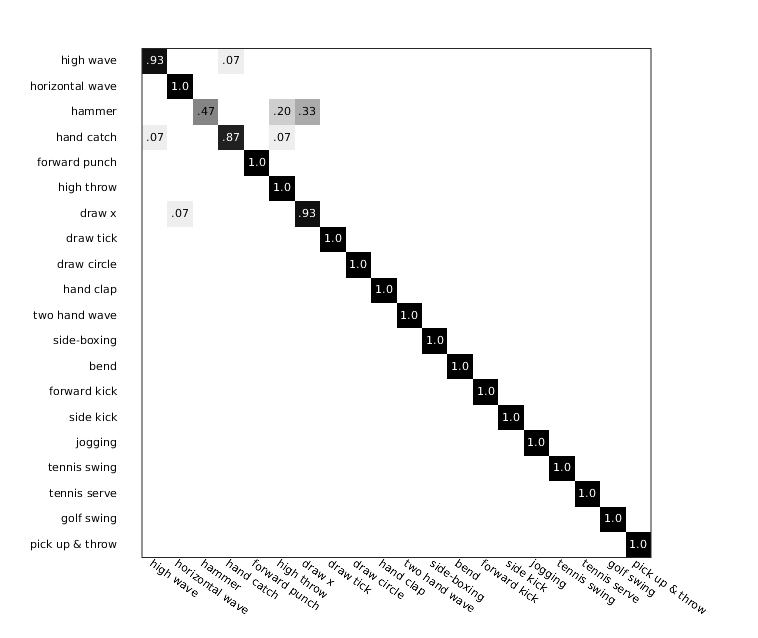}
\end{center}
   \caption{Confusion matrix of our approach on the MSR Action3D dataset: except for the \emph{hammer} class, all other action classes are classified with more than 80\% accuracy. 16 out 20 action classes are perfectly classified.}
	\label{fig: cm_action}
\end{figure*}

In Table~\ref{tab: action}, a more challenging protocol of \cite{Actionlet12} is used. Here the model is trained and tested over all 20 action classes. The results show that our method still obtains a highly accurate recognition rate, outperforming the current best state-of-the-art by a margin of 4.2\%. The confusion matrix of our method on this dataset under the second protocol is displayed in Figure~\ref{fig: cm_action}, where 16 of 20 action classes are perfectly classified. The only highly misclassified class is \emph{hammer}, because its distinctive pattern involves human-object interaction, which is not captured by the skeleton data.

\subsection{MSR DailyActivity3D}
In MSR DailyActivity3D, there are 16 action classes performed by 20 subjects twice, making up 320 action instances. Each subject performs an action class in two variants (\eg sitting versus standing, or in front of versus behind an object). This dataset has longer sequences, usually in 100-300 frames. We still follow the cross-subject test in protocol of \cite{Actionlet12} and \cite{MovingPose}, where training and testing are conducted over all action classes. Because this dataset contains more local information than MSR Action3D, we construct $H_t$ with top responding $N_A=50$ trajectorylets, select $M_A=15$ best detectors for each training instance, and reduce the final number of clustered detectors to $K=500$.

\begin{table}[h]

\begin{center}
\begin{tabular}{|l|c|}
\hline
Methods & Accuracy  \\
\hline\hline
Dynamic Temporal Warping \cite{DTW06} & $54.0$ \\
Actionlet Ensemble (skeleton data only) \cite{Actionlet12} & $68.0$ \\
Moving Pose \cite{MovingPose} \footnotemark[4] & $70.6$  \\
Ours & $\mathbf{75.0}$\\
\hline
\end{tabular}
\end{center}
\caption{Results on the MSR DailyActivity3D dataset.}
\label{tab: activity}
\end{table}
 \footnotetext[4]{Although the reported result in \cite{MovingPose} is 73.8\%,  we never achieved this accuracy with their code due to environmental factors. For a fair comparison, we used the result 70.6\%, which is the best performance under the same environment and setting with our approach.}

\begin{figure}[t]

\begin{center}
   \includegraphics[width=1.05\linewidth]{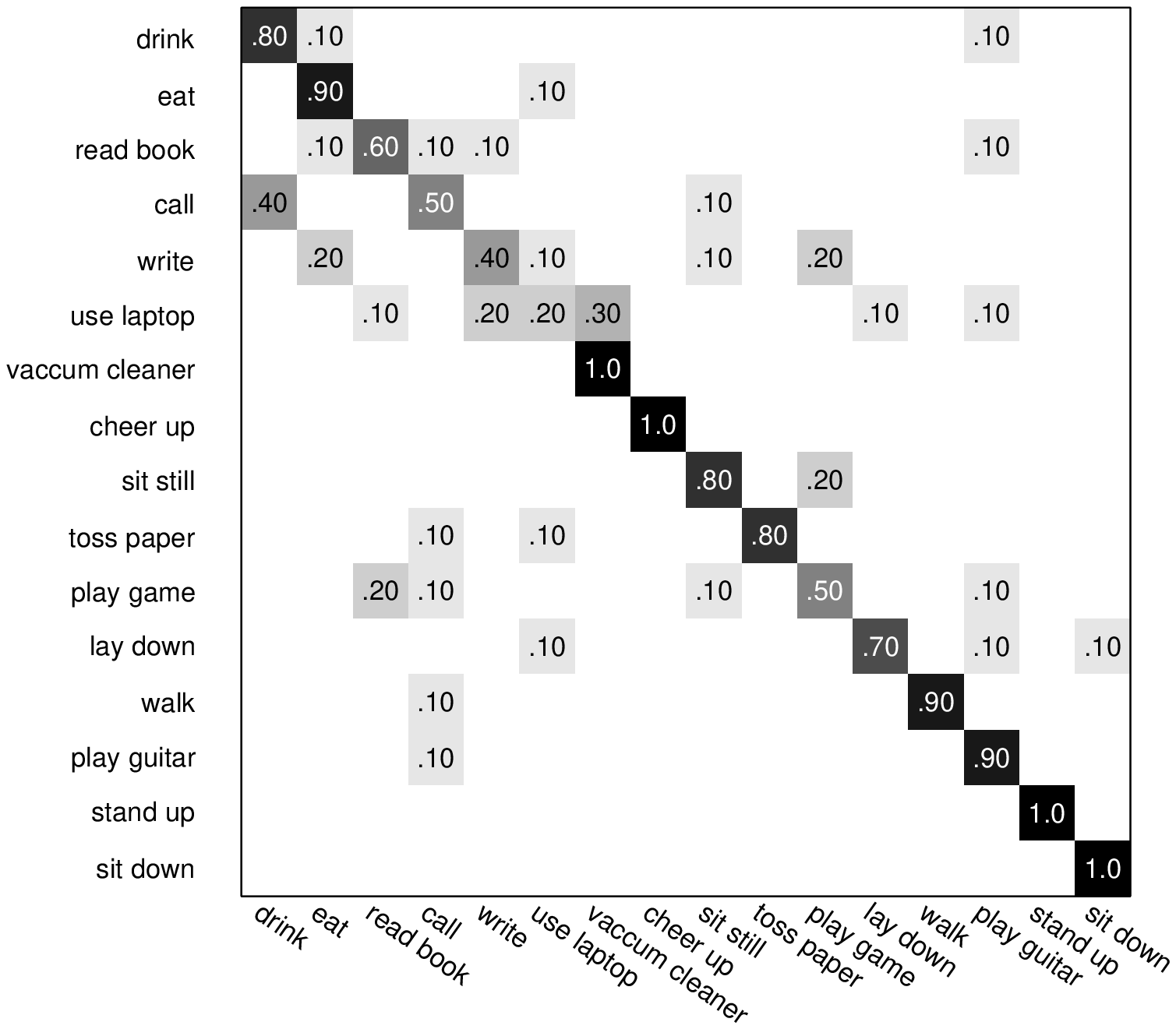}
\end{center}
   \caption{Confusion matrix of our approach on the MSR DailyActivity3D dataset: although this is a challenging dataset for skeleton-based action recognition, 11 out of 16 classes are classified with more 70\% accuracy.}
\label{fig: cm_activity}
\end{figure}

We compare our approach with other state-of-the-art methods in Table~\ref{tab: activity}. As the purpose of this experiment is to address skeleton-based action recognition, some best reported results \cite{HON4D, Actionlet12} on this dataset using additional RGB-D data are not comparable to our method, and therefore we cite the result of \cite{Actionlet12} using only skeleton data.  Although MSR DailyActivity3D share the same data structure as the MSR Action3D, it is much more challenging because: 1) the activities are complex combinations of multiple sub-actions, 2) human-object interaction information is not available in skeleton data, 3) partial occlusion by interacting objects causes the skeleton data to be highly noisy. However, the results show that our approach still outperforms all other state-of-the-art methods.  As shown in Figure~\ref{fig: cm_activity}, most of the poorly classified actions involves interaction with objects, such as \emph{read book}, \emph{call cellphone}, and \emph{use laptop}. On the other hand, non-interactive action classes like \emph{cheer up}, \emph{walk}, and \emph{sit down}, are recognized with high accuracy. This demonstrates that our method is able to capture distinctive patterns of actions in terms of ``movement'', but may be confused if some actions share similar ``movement'' patterns despite the presences of different interacting objects, because they are not described in the skeleton data.

\subsection{Parameter analysis}
\begin{figure}[h]
\begin{center}
   \includegraphics[width=1.05\linewidth]{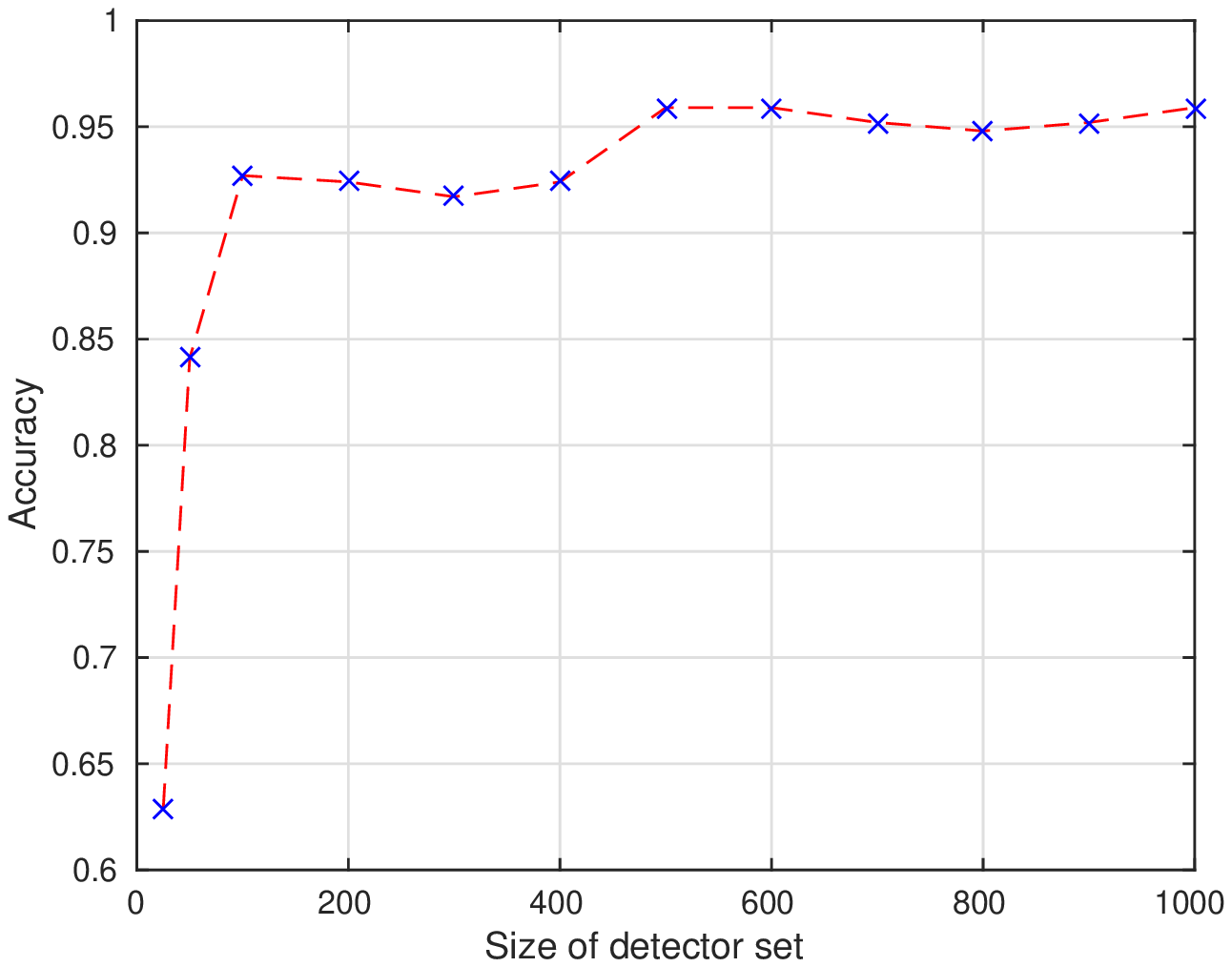}
\end{center}
   \caption{Recognition accuracies obtained from varying $K$ on the MSR Action 3D dataset: when $K \geq 500$ the results become  stable.}
	\label{fig: K_action}
\end{figure}

In this section we analyse how the parameter settings affect the performance. Using the same protocol of \cite{Actionlet12}, we provide results of MSR Action3D dataset from other parameter settings. Figure \ref{fig: K_action} illustrates the performances of our method as $K$ ranges from \{25, 50, 100, 200, 300, ... , 1000\}, while keeping  $N_A=50$ and $M_A=10$. When we set the size of detector set more than 500, the results tend to converge to a value above 94.5\%. Table \ref{tab: compare_N_M_action} presents results of choosing different pairs of $M_A$ and $N_A$ while keeping $K=500$.

\begin{table}
\begin{center}
\begin{tabular}{|l|c|c|c|c|c|c|}
\hline
\theadfont\diagbox[width=6em]{$M_A$}{$N_A$} & \thead{5}  & \thead{10} & \thead{15} & \thead{20} & \thead{30} & \thead{50} \\
\hline\hline
5 & 91.7  &  & & & & \\
10 & 92.7 & 93.4 & & & & \\
20 & 93.1 & 93.1 &94.1 &94.8 & & \\
30  & 94.8  & 95.2 & 94.1 & 93.8 & 94.8 & \\
50  & 95.5  & 95.9  & 95.9  &94.8  & 94.8 & 94.2\\

\hline
\end{tabular}
\end{center}
\caption{Results from different pairs of the $M_A$ and $N_A$ on MSR Action3D: we can obtain the best performance from multiple choices.}
\label{tab: compare_N_M_action}
\end{table}

\begin{figure}[h]
\begin{center}
   \includegraphics[width=1.05\linewidth]{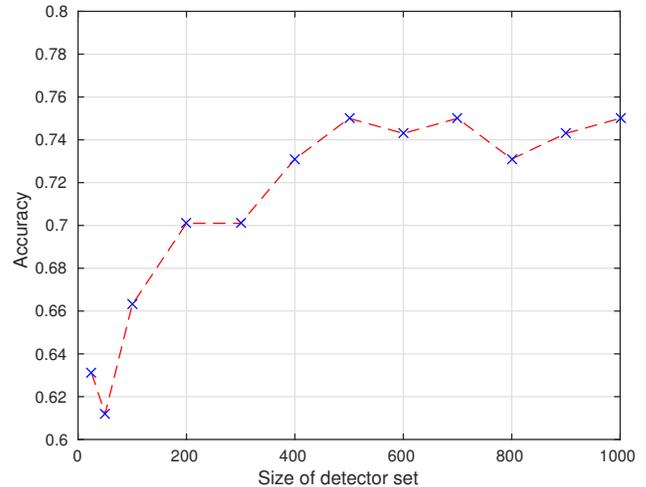}
\end{center}
   \caption{Recognition accuracy obtained from varying $K$ on the MSR Daily Activity 3D dataset: when $K>400$ the results become stable.}
	\label{fig: K_activity}
\end{figure}

For the MSR DailyActivity3D dataset, Figure \ref{fig: K_activity} illustrates the performances of our method as K ranges from \{25, 50, 100, 200, 300, ... , 1000\}, while keeping  $N_A=50$ and $M_A=15$. When $K$ is set to more than 500, the results become stable.  The effect of choosing different pairs of $M_A$ and $N_A$ is listed in in Table \ref{tab: compare_N_M_activity}. When $M_A$ is large enough, the results variation becomes small. It can be observed that, on both datasets, there are multiple choices of parameters that are able to produce the optimal result and this verifies the robustness of our approach.

Table \ref{tab: TP_levels} shows the results under different temporal pyramid settings for the two datasets. A typical 3-level pyramid is the best choice for MSR DailyActivity3D as low level pyramids fail to grasp the temporal information while higher level ones brings too much noise. On the other hand, when temporal pyramid is applied to MSR Action3D, the performance is worsened.

\begin{table}
\begin{center}
\begin{tabular}{|l|c|c|c|c|c|c|}
\hline
\theadfont\diagbox[width=6em]{$M_A$}{$N_A$} & \thead{5}  & \thead{10} & \thead{15} & \thead{20} & \thead{30} & \thead{50} \\
\hline\hline
5 & 68.7  &  & & & & \\
10 & 68.1 & 69.4 & & & & \\
20 & 68.7 & 71.2 & 70.0 & 69.4 & & \\
30  & 70.0 & 73.1 & 73.8 &71.2 & 71.2 & \\
50  & 73.1 & 74.3 & 75.0 &75.0 & 74.3 & 71.9\\

\hline
\end{tabular}
\end{center}
\caption{Results from different pairs of the $M_A$ and $N_A$ on MSR DailyActivity3D.}
\label{tab: compare_N_M_activity}
\end{table}

\begin{table}
\begin{center}
\begin{tabular}{|l|c|c|c|c|}
\hline
TP level & 1  & 2 & 3 & 4\\
\hline\hline
Action & 95.9  & 92.4 & 89.7  & N/A\\
DailyActivity & 66.3  & 70.6 & 75.0  & 68.8\\

\hline
\end{tabular}
\end{center}
\caption{Results obtained from different temporal pyramid levels on MSR Action3D and MSR DailyActivity3D datasets.}
\label{tab: TP_levels}
\end{table}

\subsection{Power of local trajectorylet descriptor}
The moving pose descriptor proposed in \cite{MovingPose} captures local information at frame-level of human skeleton actions. Our trajectorylet can be seen as a natural extension of it in the sense that we extend the dynamic information from frame-level range to a longer temporal range. In order to demonstrate the power of our descriptor we now apply our template detector learning framework to moving pose descriptor and compare its performance with that of trajectorylet.

In order to evaluate the effect of varying $L$ on performance we have varied the length our trajectorylet from (3, 5, 7). Table \ref{tab: compare_feature} shows that using the same detector learning and classification approach, trajectorylets achieve better results on both datasets for all tested values of $L$. As seen, this extension of moving pose descriptor is superior over the original design. It is worth noting that performance does not necessarily improve as the length of trajectorylets increases. A moderate length of trajectorylet ($L=5$) leads to the best performance.

We also test the effect of using different components of the trajectorylet descriptor. In our experiment, we examine the performance of single dynamic components, including static pose $\bx_0$, relative joint displacement $\bx_1$, velocity $\bx_2$, and their combinations. We also further define an acceleration component analogous to \eqref{EQ:adjacent_frame1} and \eqref{EQ:adjacent_frame2}:
\begin{align}
\label{EQ:Com_acc}
 & \bx_{3}^{t_0} = [ {\Delta^3\jointPos^{t_0+2}}^\T, \cdots, {\Delta^3\jointPos^{t_0+L-1}}^\T ]^\T \in  \mathbb{R}^{((L-3) \times 3J)}, \\
 &
\Delta^3\jointPos^{t_0+i}=\Delta^2\jointPos^{t_0+i} - \Delta^2\jointPos^{t_0+i-1}, \, i=3, \cdots, L-1. \notag
\end{align}

 The results of varying settings of a trajectorylet with $L=5$ are listed in Table \ref{tab: compare_level}.  We find that the dynamic components of $\bx_1$ and $\bx_2$ alone do not show promising results. However, when combined with static $\bx_0$, the performance is significantly improved. Table \ref{tab: compare_level} also shows that the additional acceleration component in \eqref{EQ:Com_acc} does not improve the performance.

\begin{figure*}[h]
\vspace{-0.0cm}
\begin{center}
\scalebox{0.6}{
\begin{tabular}{@{}c@{}c@{}c@{}c@{}c}
\includegraphics[width=0.18\linewidth]{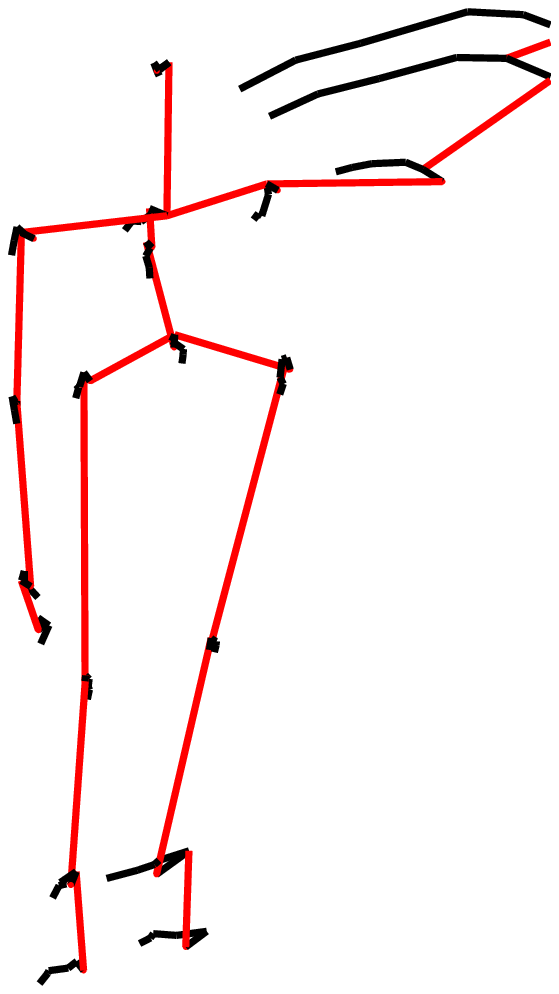} \ &
\includegraphics[width=0.18\linewidth]{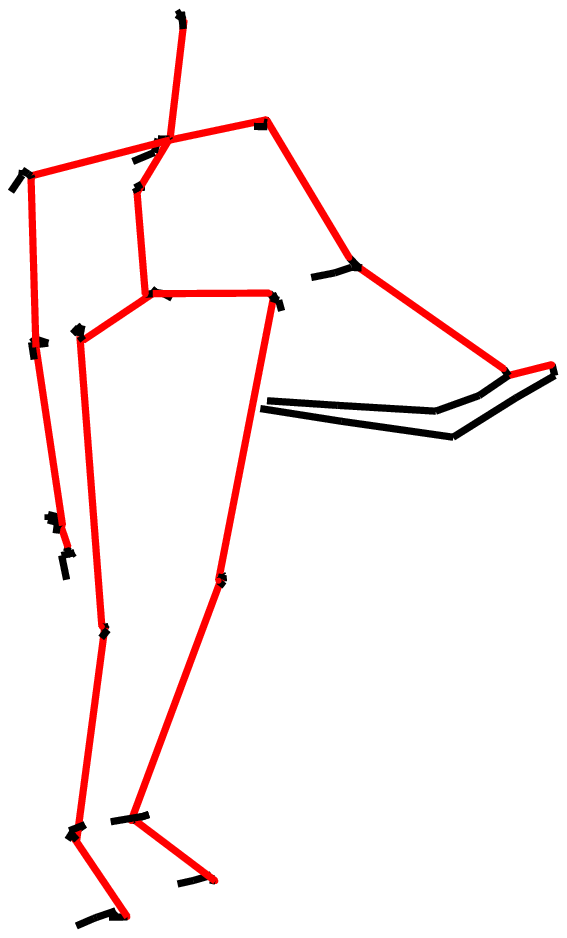} \ &
\includegraphics[width=0.15\linewidth]{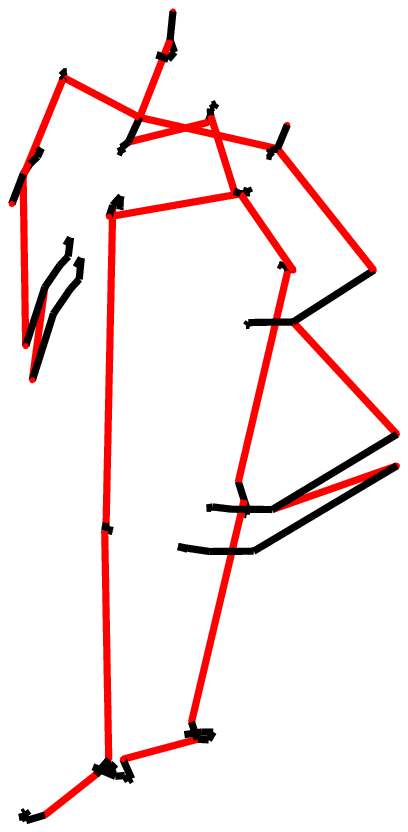} \ &
\includegraphics[width=0.16\linewidth]{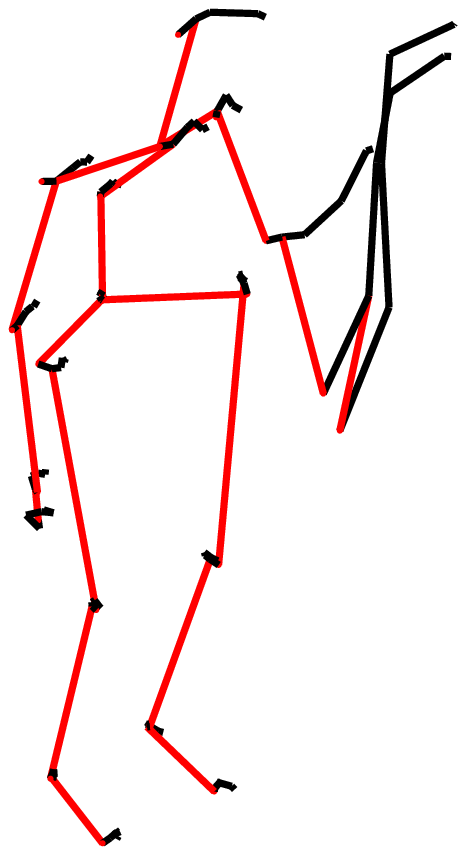} \ &
\includegraphics[width=0.18\linewidth]{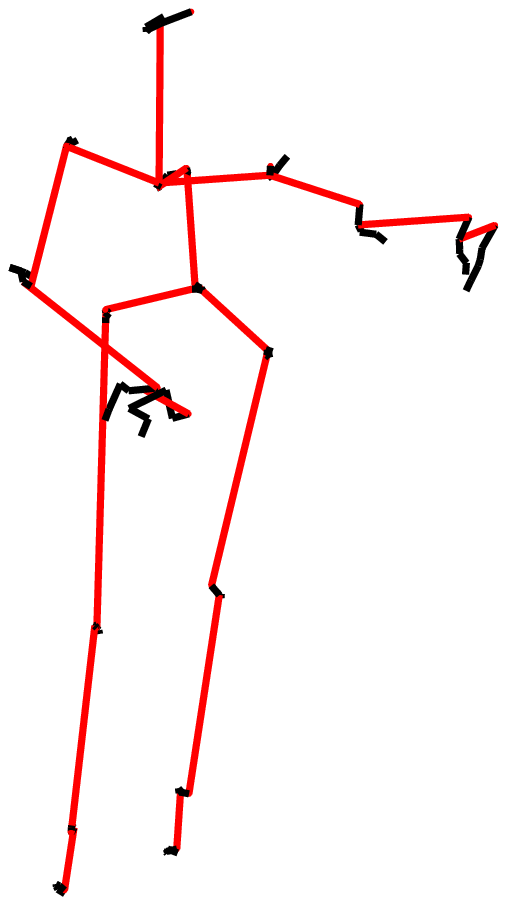} \\
high wave\ &
hori. wave \ &
hammer \ &
hand catch \ &
forward punch\\
\includegraphics[width=0.13\linewidth]{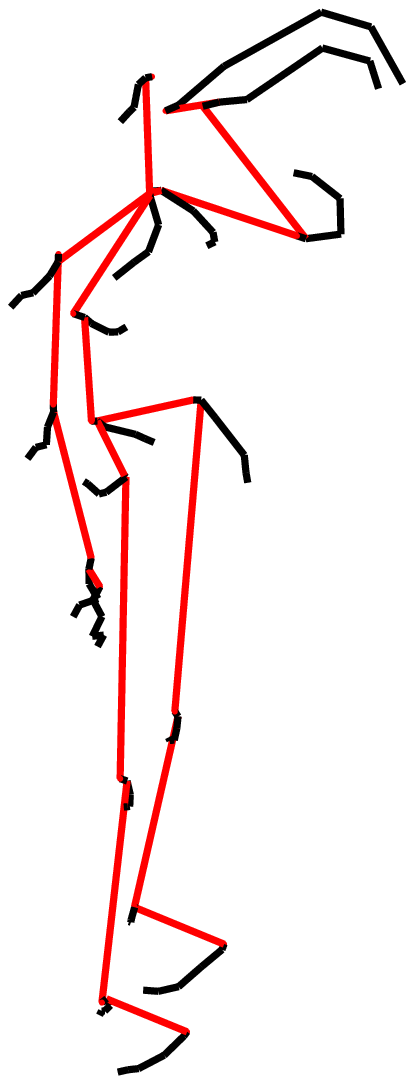} \ &
\includegraphics[width=0.18\linewidth]{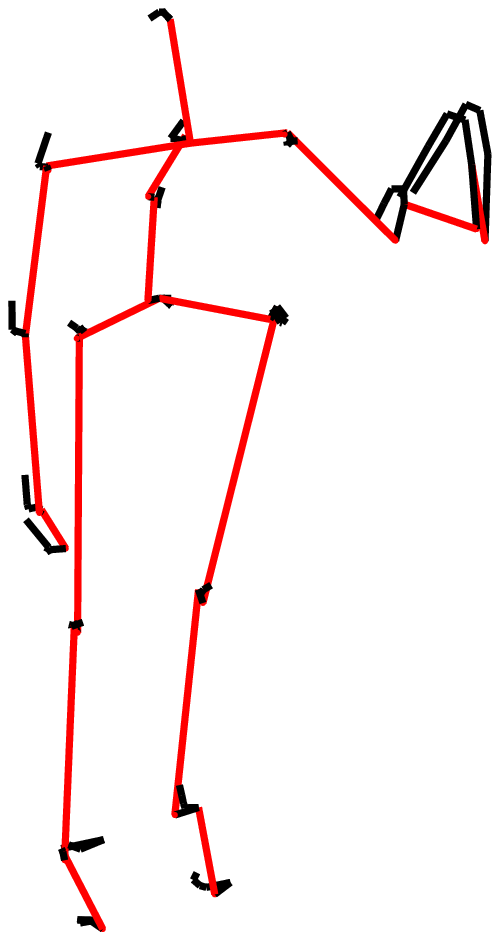} \ &
\includegraphics[width=0.16\linewidth]{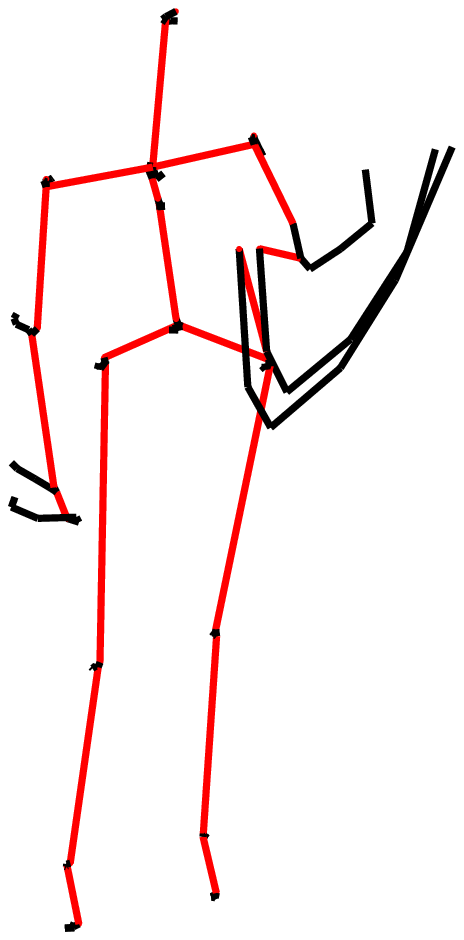} \ &
\includegraphics[width=0.16\linewidth]{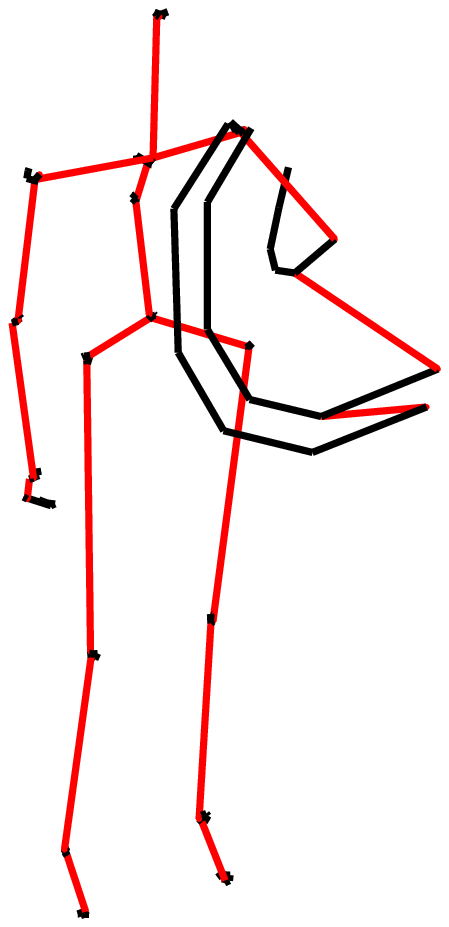} \ &
\includegraphics[width=0.21\linewidth]{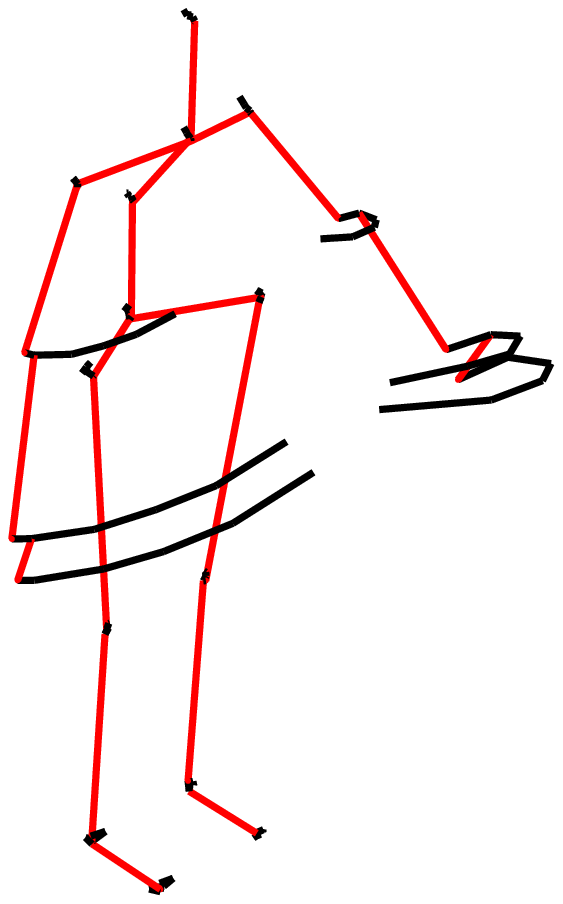} \\
high throw\ &
draw x \ &
draw tick \ &
draw circle \ &
hand clap\\
\includegraphics[width=0.21\linewidth]{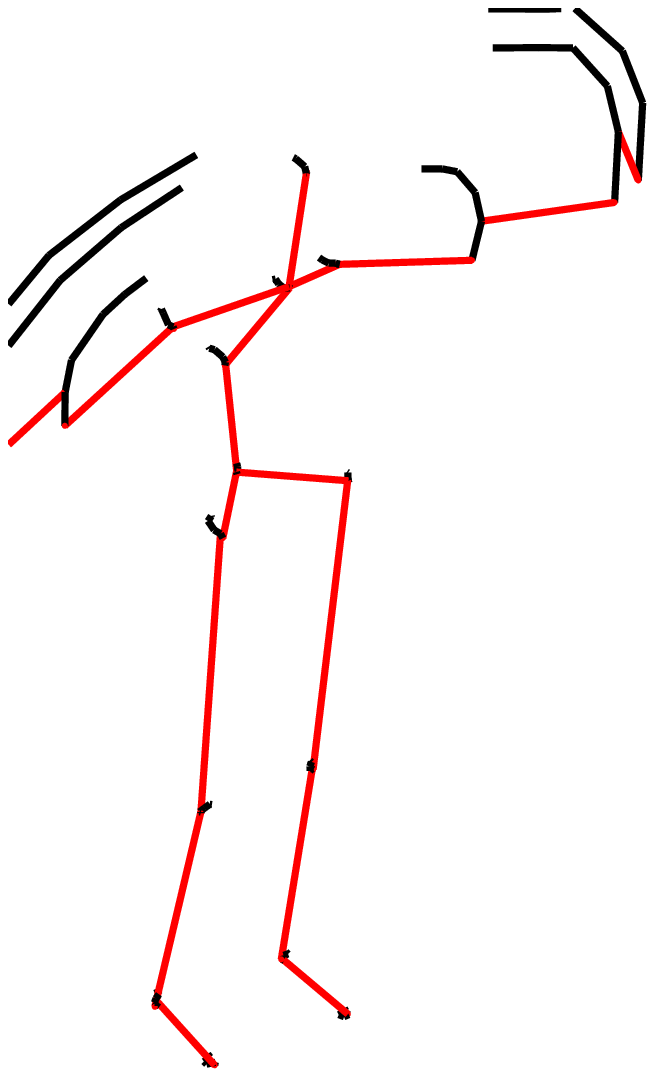} \ &
\includegraphics[width=0.18\linewidth]{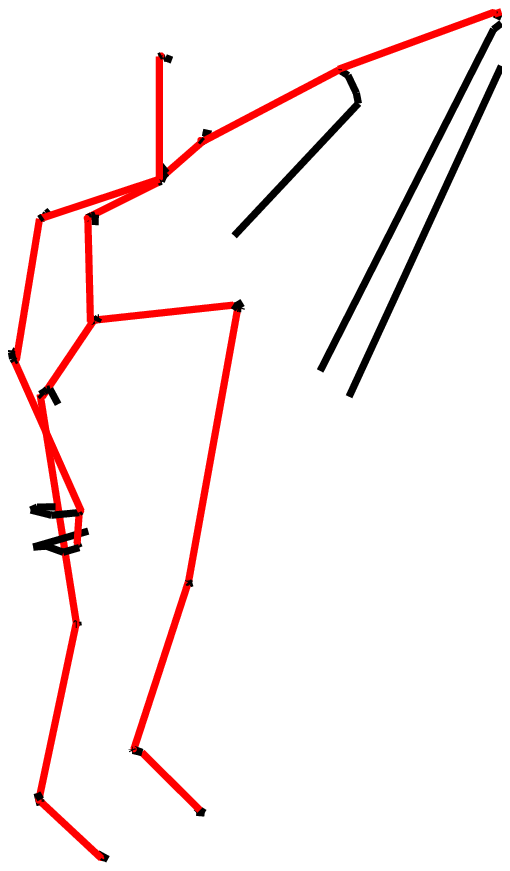} \ &
\includegraphics[width=0.11\linewidth]{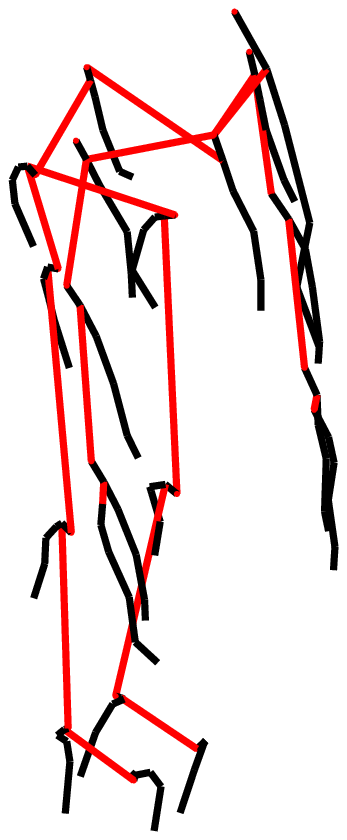} \ &
\includegraphics[width=0.15\linewidth]{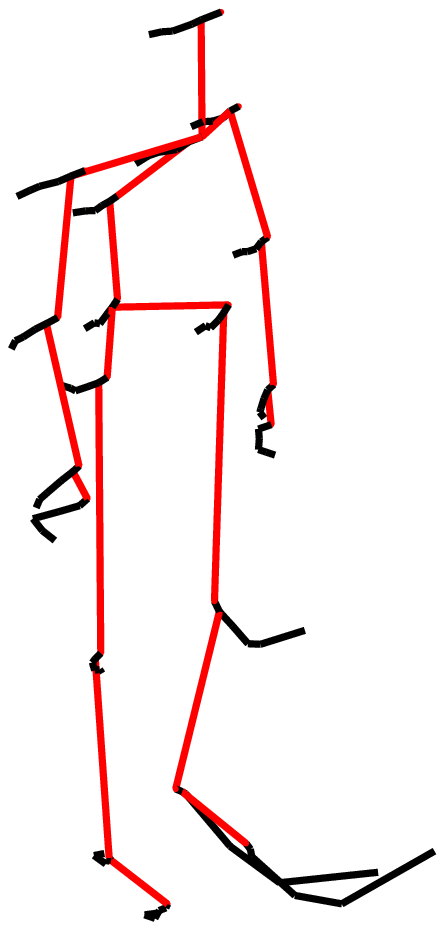} \ &
\includegraphics[width=0.20\linewidth]{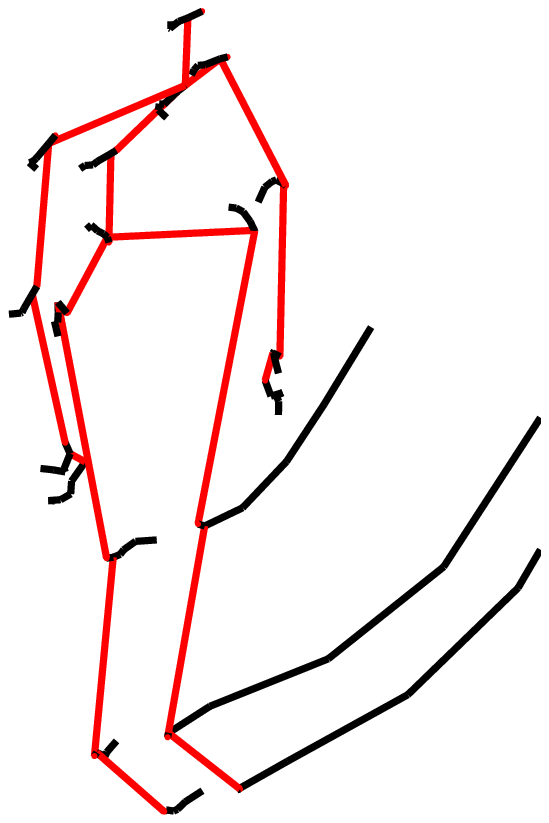} \\
2-hand wave\ &
boxing \ &
bend \ &
forward kick \ &
side kick\\
\includegraphics[width=0.12\linewidth]{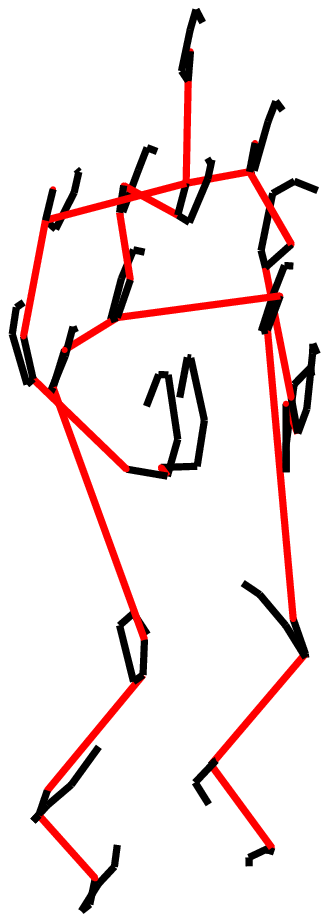} \ &
\includegraphics[width=0.19\linewidth]{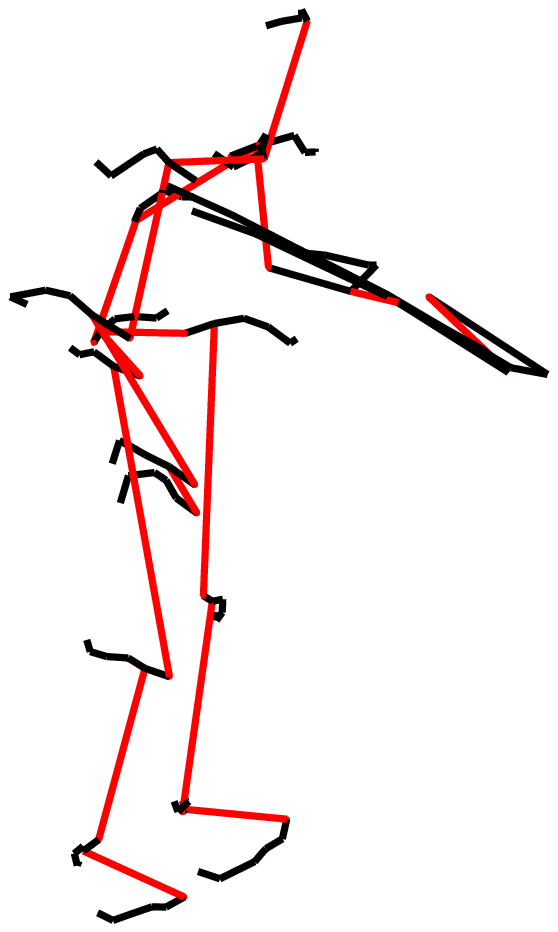} \ &
\includegraphics[width=0.13\linewidth]{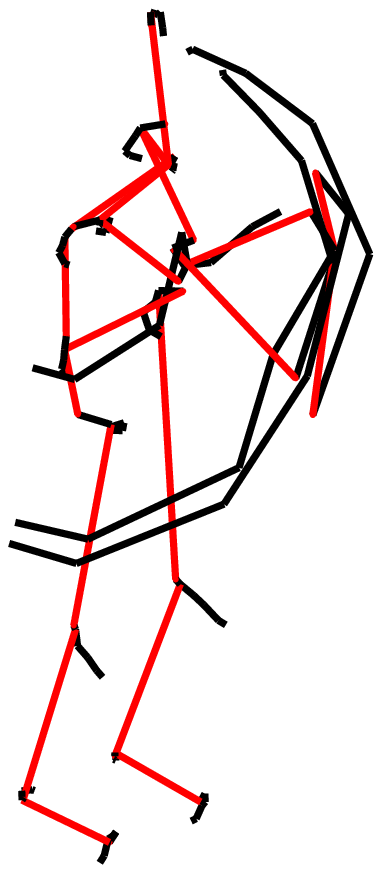} \ &
\includegraphics[width=0.15\linewidth]{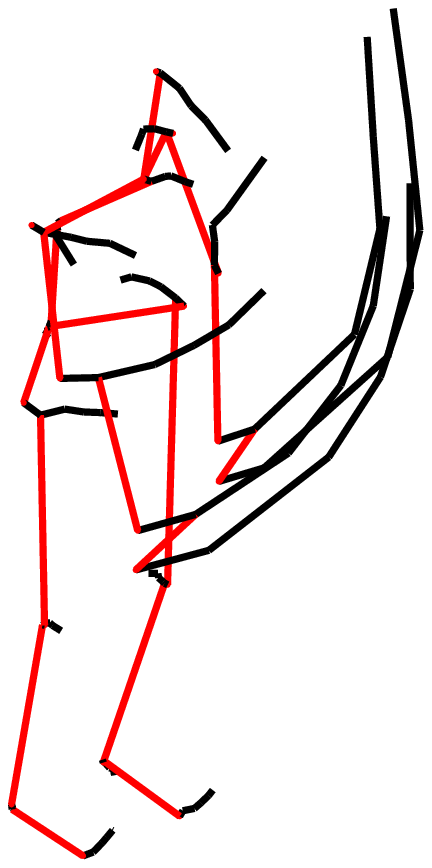} \ &
\includegraphics[width=0.17\linewidth]{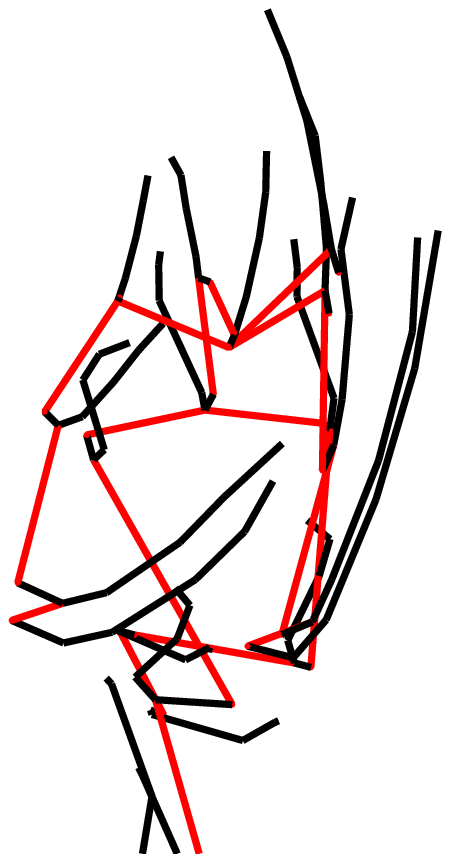} \\

jog\ &
t. swing \ &
t. serve \ &
g. swing \ &
pick up \& throw\\

\end{tabular}
}
\end{center}
\caption{Some examples responding on the template detector set of MSR Action3D. The black curves represent current trajectorylets of the red skeleton. The fact that the our approach identifies discriminative patterns of movement seems clear.}
\label{fig:rep_trajectorylets}
\end{figure*}

\begin{table}[ht]
\begin{center}
\begin{tabular}{|l|c|c|}
\hline
Descriptors & MSR Action   & MSR DailyActivity\\
\hline\hline
Moving Pose \cite{MovingPose} & 91.7  & 71.3\\
Ours($L=3$) & 93.1 & 72.5\\
Ours($L=5$) & $\mathbf{95.9}$ & $\mathbf{75.0}$\\
Ours($L=7$) & $\mathbf{95.9}$ & 73.1\\

\hline
\end{tabular}
\end{center}
\caption{Comparison of using different descriptors.}
\label{tab: compare_feature}
\end{table}

\begin{table}[ht]
\begin{center}
\begin{tabular}{|l|c|c|}
\hline
Component & MSR Action   & MSR DailyActivity\\
\hline\hline
$\bx_0$ & 92.4 & 72.5\\
$\bx_1$ & 91.7 & 50.3\\
$\bx_2$ & 90.3 & 42.5\\
$(\bx_0, \bx_1)$ & 93.8 & 73.1\\
$(\bx_0, \bx_1, \bx_2)$ & $\mathbf{95.9}$ & $\mathbf{75.0}$ \\
$(\bx_0, \bx_1, \bx_2, \bx_3)$ & $\mathbf{95.9}$  & 74.3 \\

\hline
\end{tabular}
\end{center}
\caption{Comparison of different using different components of trajectorylet ($L=5$).}
\label{tab: compare_level}
\end{table}

\begin{table}[ht]
\begin{center}
\begin{tabular}{|l|c|c|}
\hline
Method & MSR Action   & MSR DailyActivity\\
\hline\hline
VLAD \cite{VLAD10} & $83.1 $  & $51.9$\\
LLC \cite{LLC10} & $90.7 $ & $65.6 $\\
LSC \cite{LSC11} & $92.1 $ & $66.9 $\\
Ours & $\mathbf{95.9}$ & $\mathbf{75.0 }$\\
\hline
\end{tabular}
\end{center}
\caption{Comparison of feature learning methods.}
\label{tab: learning_compare}
\end{table}

\subsection{Power of template detector learning}

Our method generates action representation from learned detector set of discriminative trajectorylets. In this section, we compare this method with three state-of-the-art bag-of-feature techniques that learn middle-level feature from the same local trajectorylet feature: VLAD (vector of locally aggregated descriptors)\cite{VLAD10}, LLC (locality-constrained linear coding)\cite{LLC10}, and LSC (localized soft-assignment coding) \cite{LSC11}.

We train codebook of the same size $K=128$ with k-means for all three methods, and set the neighbourhood size of codewords as $\kappa=10$ for LSC and LLC. The results listed in Table~\ref{tab: learning_compare} show, for the task of action recognition, our proposed feature learning framework produces the most discriminative action representation, compared with the state-of-the-art methods. Figure~\ref{fig:rep_trajectorylets} illustrates some  trajectorylets fired on the template detector set of MSR Action3D. It is clear that they show representative patterns for the corresponding action classes.

\section{Conclusion}
\label{sec: CONC}
This work describes an effective skeleton-based action approach that achieves high accuracy on the relevant benchmark datasets. The keys to this performance are two factors. We propose trajectorylet, a novel local descriptor that captures static and dynamic information in a short interval of joint trajectories. We also devise a novel framework to generate robust and discriminative representation for action instances by learning a set of distinctive trajectorylet detectors. On two benchmark datasets acquired from the Kinect sensor, our method outperforms, to our knowledge, all existing approaches by a significant margin. We also separately demonstrate the validity of our local descriptors and template detector learning method. To further expand our framework, we plan to incorporate local temporal information to enable real-time detection, as well as investigate the RGB data to study the involvement of human-object interactions.

\ifCLASSOPTIONcaptionsoff
  \newpage
\fi

{\small
\bibliographystyle{ieee}
\bibliography{CSRef}
}

\end{document}